\documentclass{article}

\PassOptionsToPackage{numbers, compress}{natbib}
\usepackage[final]{nips_dlps_2017}
\usepackage[utf8]{inputenc} 
\usepackage[T1]{fontenc}    
\usepackage[hidelinks]{hyperref}       
\usepackage{url}            
\usepackage{booktabs}       
\usepackage{amsfonts}       
\usepackage{nicefrac}       
\usepackage{microtype}      
\usepackage{todonotes}
\usepackage{subfigure}
\usepackage{listings}
\usepackage{adjustbox} 

\usepackage{mathtools}
\usepackage{graphicx}
\usepackage[nameinlink,noabbrev]{cleveref}
\usepackage{xcolor}

\newcommand{\CPProb}{\textsc{CPProb}}       
\newcommand{\SHERPA}{\textsc{SHERPA}}       
\newcommand{\GEANT}{\textsc{GEANT}}        

\newcommand{\code}{\texttt}                 

\let\epsilon\varepsilon                     

\makeatletter
\newcommand*\ie{\textit{i.e.}\@ifnextchar.{\@gobble}{\relax}}
\newcommand*\vs{\textit{vs.}\@ifnextchar.{\@gobble}{\relax}}
\newcommand*\etc{\textit{etc.}\@ifnextchar.{\@gobble}{\relax}}
\newcommand*\eg{\textit{e.g.}\@ifnextchar.{\@gobble}{\relax}}
\makeatother

\definecolor{listinggray}{gray}{0.9}
\title{Improvements to Inference Compilation for Probabilistic Programming in \\Large-Scale Scientific Simulators}

\author{
  Mario Lezcano Casado, Atılım Güneş Baydin\\\textbf{David Martínez Rubio, Tuan Anh Le, Frank Wood} \\
  Department of Engineering Science\\
  University of Oxford\\
  \texttt{\{lezcano,gunes,damaru,tuananh,fwood\}@robots.ox.ac.uk} \\
\And
  Lukas Heinrich, Gilles Louppe, Kyle Cranmer \\
  Department of Physics \& Center for Data Science\\
  New York University \\
  \texttt{\{cranmer,lukas.heinrich,g.louppe\}@cern.ch} \\
\And
  Karen Ng \\
  Intel Corporation\\
  \texttt{karen.y.ng@intel.com}
\And
  Wahid Bhimji, Prabhat \\
  Lawrence Berkeley National Laboratory \\
  \texttt{\{wbhimji,prabhat\}@lbl.gov}\\
}

\begin{document}

\maketitle

\begin{abstract}
We consider the problem of Bayesian inference in the family of probabilistic models implicitly defined by stochastic generative models of data. In scientific fields ranging from population biology to cosmology, low-level mechanistic components are composed to create complex generative models. These models lead to intractable likelihoods and are typically non-differentiable, which poses challenges for traditional approaches to inference. We extend previous work in ``inference compilation'', which combines universal probabilistic programming and deep learning methods, to large-scale scientific simulators, and introduce a C++ based probabilistic programming library called \CPProb{}. We successfully use \CPProb{} to interface with \SHERPA{}, a large  code-base used in particle physics. Here we describe the technical innovations realized and planned for this library.
\end{abstract}

\section{Introduction}

Complex simulations are used for stochastic generative models for data across a wide segment of the scientific community, with applications as diverse as hazard analysis in seismology \citep{heinecke2014petascale},  supernova shock waves in astrophysics \citep{endeve2012turbulent}, market movements in economics \citep{reberto2001agent}, and blood flow in biology \citep{perdikaris2016multiscale}. In these generative models, complex simulations are composed from low-level mechanistic components. These models lead to intractable likelihoods and are typically non-differentiable, which renders many traditional statistical inference algorithms irrelevant and motivates a new class of so-called likelihood-free inference algorithms \citep{hartig2011statistical}.

There are two broad strategies for this type of likelihood-free inference problems. In the first, one uses a simulator indirectly to train a surrogate model endowed with a likelihood that can be used in traditional inference algorithms, for example approaches based on conditional density estimation~\cite{JMLR:v17:16-272,papamakarios2017masked, rezende2015variational, NIPS2016_6581} and density ratio estimation~\citep{cranmer2015approximating,dutta2016likelihood}. Alternatively, approximate Bayesian computation (ABC) \citep{wilkinson2013approximate,sunnaaker2013approximate} refers to a large class of approaches for sampling from the posterior distribution of these likelihood-free models where the original simulator is used directly as part of the inference engine. While variational inference algorithms are often used when the posterior is intractable, they are not directly applicable when the likelihood of the data generating process is unknown. 

The class of inference strategies that directly use a simulator avoids the necessity of approximating the generative model. Moreover, using a domain-specific simulator offers a natural pathway for inference algorithms that provide interpretable posterior samples with rich semantics. In this work, we take this approach, extend previous work in probabilistic programming \citep{gordon2014probabilistic} and inference compilation \citep{le2017inference} to large-scale complex simulators, and demonstrate the ability to instrument and control simulators written in C++, which is the language of choice for many large-scale simulators in science and industry.

Our work is primarily motivated by applications in high-energy physics (HEP), which studies elementary particles and their interactions using energetic events created in particle accelerators such as the Large Hadron Collider (LHC) at CERN. The observed data are the result of interactions of particles generated in a collision event and observed through particle detectors. From these observations, we want to infer the properties of the particles and interactions that generated them. 

Ultimately, our goal is to be able to infer the properties of the Higgs boson \citep{2012PhLB..716....1A, 2012PhLB..716...30C} using probabilistic programming with the state-of-the-art simulators \SHERPA{} \citep{Gleisberg:2008ta}, a Monte Carlo ``event generator'' of high-energy reactions of particles, and \GEANT{} \citep{allison2016recent}, a toolkit for the simulation of the passage of the resulting particles through the detector. Both simulators are implemented in C++ and encapsulate deep domain knowledge in significantly large code bases (> 1 M lines of code), providing a challenging setting for demonstrating our technique and its scalability.


Our key innovations in this paper include: (1) the interfacing of \CPProb{} with the existing \SHERPA{} code base with very few modifications; (2) the ability to inspect execution traces leading to a novel probabilistic model debugging tool; and (3) a new code annotation scheme for dealing with traces of unbounded length in algorithms such as rejection sampling encountered in HEP settings.

\section{CPProb}


In order to equip large-scale simulation code bases with inference compilation, we developed \CPProb{} \citep{lezcano2017cpprob},\footnote{Code will be publicly released at a future time.} a probabilistic programming library written in C++14 that allows inference in probabilistic models written in C++. Using a universal probabilistic programming approach means that our library enables inference in existing simulator codes without having to rewrite the model in a specific probabilistic programming language. This is done simply by redirecting calls of the random number generation and introducing annotations to the code defined by \CPProb{}.

The library exports three main functions: \code{sample}, \code{observe}, and \code{predict}, all of which are statically typed both for efficiency and to offer compile-time type guarantees related to the addressing scheme used in \CPProb{}. The \code{sample} and \code{observe} statements allow \CPProb{} to assign addresses to and keep track of probabilistic execution traces of the simulator during training and inference. Addresses identify random choices according to their structural position in the execution, similar to a stack trace \citep{wingate2011lightweight}. A \code{sample} statement generates a random value from a specified probability distribution, and appends this sample to the execution trace. An \code{observe} statement specifies the conditioning of a random variable upon an observed value, such as the conditioning of the simulation output on data. A \code{predict} statement is used to designate any latent values within the simulation that we would like to be reported as the result of inference.

The main algorithm used in \CPProb{} is importance sampling \citep{doucet2009tutorial} in conjunction with inference compilation~\cite{le2017inference}. The inference compilation approach combines universal probabilistic programming and deep learning methods so that an \emph{inference network} is trained to choose the parameters of a family of proposal distributions in a sequential importance sampling (SIS) inference engine. The inference network controls the execution of the simulator and provides substantially more efficient sampling for ABC than Markov chain Monte Carlo (MCMC) methods. Since SIS reaches deep into the control flow of the program, the posterior samples are efficiently obtained.

\textbf{Interpretability:} The ability to connect posterior samples  to the simulator code is a key advantage of our method. This connection enables deep interpretability of inference results in the context of the physically-motivated latent process encoded by the simulator. Such interpretability is crucial for applications in the physical sciences. For instance, this approach gives us the ability to inspect any aspect of the latent process encoded in the simulation, such as the chain of particle decays and interactions within the detector that led to particular posterior predictions. This capability is not  possible in inference techniques that do not have access to the simulator, such as those based on neural networks.

\begin{figure}[!t]
\centering
\includegraphics[width=0.35\textwidth]{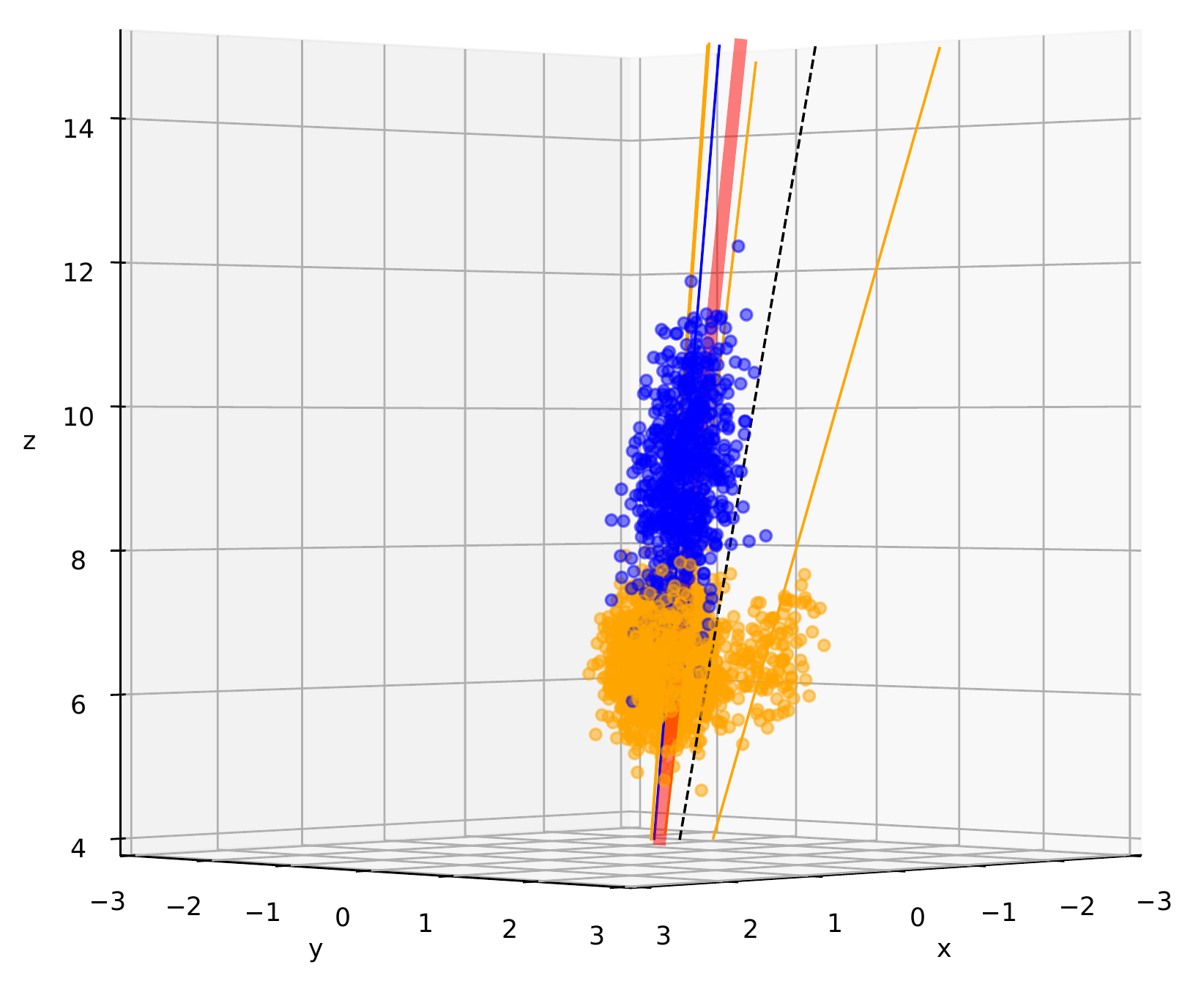}
\includegraphics[width=0.358\textwidth]{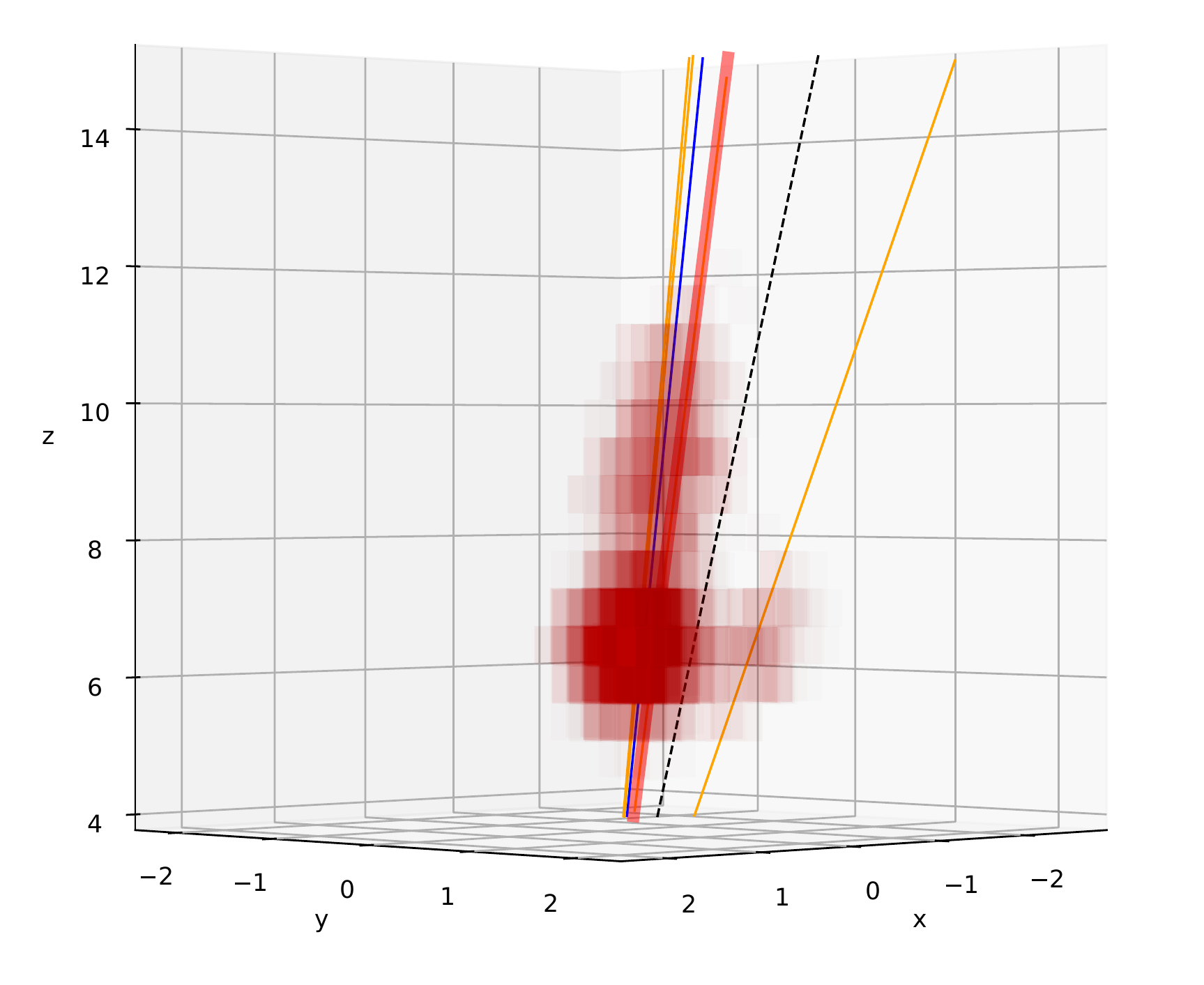}
\caption{Simulated $\tau$ lepton decay event in the channel $\tau \to \nu_\tau \pi^- \gamma \gamma$. The red line is the $\tau$ momentum vector. \emph{Left:} The latent momentum vectors (lines) and interactions in the calorimeter (dots) for the interacting $\tau$ decay products. \emph{Right:} The observed energy deposits (with latent momentum vectors overlaid).}
\label{fig:sherpa_heat_trace}
\vspace{-.1in}
\end{figure}

\section{Results}

We focus our initial studies on the decay of $\tau$ leptons, whose subsequent decay products interact in the detector. 
This provides a rich and realistic  particle physics use-case, and directly connects to the physics of the Higgs boson. During simulation, \SHERPA{} stochastically selects a set of particles to which the initial $\tau$ lepton will decay---a ``decay channel''---and samples the momenta of these particles according to a joint density obtained from underlying physical theory. 
Those particles will then interact in the detector leading to observations in the raw sensor data. While \GEANT{} is typically used to model the interactions in detector, for our initial studies we created a fast, approximate detector simulation for a calorimeter with longitudinal and transverse segmentation (20x35x35). The fast detector simulation deposits most of the energy for electrons and $\pi^0$ into the first layers and charged hadrons (e.g., $\pi^\pm$)  deeper into the calorimeter with larger fluctuations.
Given an observation such as the one presented in Figure~\ref{fig:sherpa_heat_trace}, we would like to infer the initial momenta of the $\tau$ lepton and the decay channel that it followed. We were able to successfully interface \CPProb{} to \SHERPA{} with minimal code annotations and successfully train an inference network artifact capable of performing inference on simulated test data. In order to achieve this, and to enhance the value of inference in this domain, it was necessary to develop several new features described in the rest of this section.



\begin{figure}[!t]
\centering
\subfigure[\hspace{46.6mm} (a) Address succession graph.]{
\includegraphics[width=0.8\textwidth]{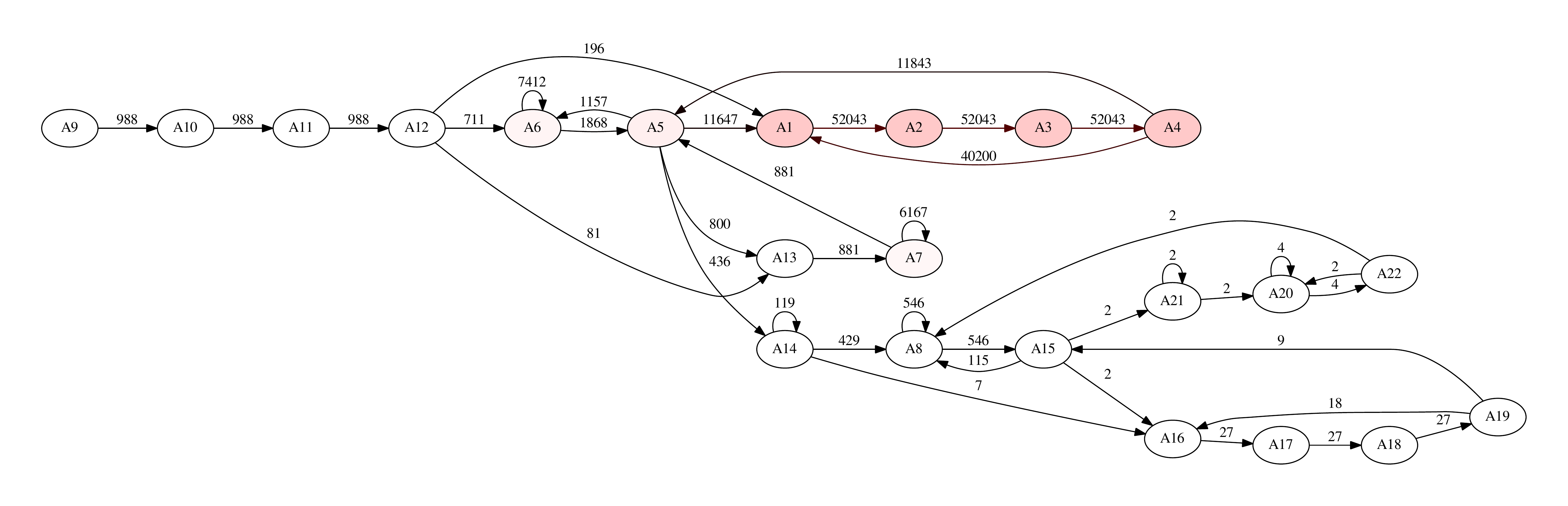}
}\vspace{-1.6cm}

\subfigure[\tiny\texttt{SHERPA-MC-2.2.3/PHASIC++/Channels/Multi-Channel.C}]{
\raisebox{-3.7mm}{\adjustbox{trim=-3mm -2mm 0 -2mm,clip}{\resizebox{0.43\textwidth}{!}{
\begin{lstlisting}[firstnumber=325]
double rn  = ran->Get();
double sum = 0;
for (size_t i=0;;++i) {
  if (i==channels.size()) {
    rn  = ran->Get();
    i   = 0;
    sum = 0.;
  }
  sum += channels[i]->Alpha();
  if (sum>rn) {
    channels[i]->GeneratePoint(p,cuts);
    if (nin==2) for (int i(0);i<nin+nout;++i) cms.BoostBack(p[i]);
    m_lastdice = i;
    break;
  }
}  
\end{lstlisting}
}}}}
\subfigure[\tiny\texttt{SHERPA-MC-2.2.3/PHASIC++/Channels/Rambo.C}]{
\raisebox{-11mm}{\adjustbox{trim=-3mm -2mm 0 -2mm,clip}{\resizebox{0.5\textwidth}{!}{
\begin{lstlisting}[firstnumber=72]
for(i=nin;i<nin+nout;i++) {
  C     = 2*ran->Get()-1;
  S     = sqrt(1-C*C);
  F     = 2*M_PI*ran->Get();
  Q     = -log( std::min( 1.0-1.e-10, std::max(1.e-10,ran->Get()*ran->Get()) ) );
  p[i]  = Vec4D(Q, Q*S*::sin(F), Q*S*cos(F), Q*C);
  R    += p[i]; 
}
\end{lstlisting}
}}}}

\caption{The source of very long execution traces within the SHERPA code base pinpointed by our probabilistic model debugging tool. (a) The probabilistic structure of the $\tau$ decay simulator. (b) The A5 node maps to the ``\texttt{ran->Get()}'' call in this source file. (c) The A1 -- A4 nodes map to the four ``\texttt{ran->Get()}'' calls in this source file.}
\label{fig:rejection_sampling}
\end{figure}

\textbf{Trace inspection:} We developed a general-purpose probabilistic model debugging/inspection tool, which allows for detailed inspection and visualization of execution traces, mapping them back to underlying code (illustrated in Figure \ref{fig:rejection_sampling}).  
This feature also allows for debugging of problems with the whole chain of applying probabilistic programming to existing large-scale code bases, where the underlying functions and order of execution may be opaque.

In our initial experiments with inference compilation of the $\tau$ lepton decay problem in \SHERPA{}, we were faced with probabilistic execution traces that were too long to efficiently backpropagate through in the inference network. While the average trace length was around $250$, we encountered traces as long as $10,000$, which made training of the inference network in the manner described in \citet{le2017inference} prohibitively slow.  We were able to find the cause of these very long traces using our inspection tool, which pinpointed a rejection sampling loop in the matrix element implementation present in the code base \citep{kleiss1994weight,kleiss1986new}. In Figure~\ref{fig:rejection_sampling}(a) we present the address succession graph accumulated over 988 traces, describing the overall probabilistic structure of the $\tau$ decay simulation. The nodes indicate \code{sample} addresses in the execution that---being concatenations of stack positions at runtime---can be traced to positions in the code base. The edges indicate deterministic execution paths of the simulator labeled by the number of times a path is traced. The bottom part of Figure~\ref{fig:rejection_sampling}(b, c) shows the locations in the \SHERPA{} source code where \code{sample} statements A1 -- A5 reside. Precisely identifying these \code{sample} statements enabled us to overcome this bottleneck as described below.

\textbf{Rejection sampling scheme:} Rejection sampling sections in the simulator pose a problem for our approach, as they define execution traces that are a priori unbounded; and since the inference network has to backpropagate through every sampled value, this makes the training significantly slower. Rejection sampling is key to the application of Monte Carlo methods for evaluating matrix elements and other stages of event generation in particle physics; thus an efficient treatment of this construction is primal. We solve this problem by implementing a novel scheme where only the last instances of random choices in a rejection sampling loop are considered during training. During inference we just use the first proposal distribution for every loop execution. Intuitively, this accounts for just training the inference network with the data that makes the loop end, effectively training the network to select proposal distributions such that the rejection loop is exited in as few iterations as possible.

This scheme requires annotating the simulator code to demarcate the parts where it should be applied, but this was greatly simplified as we were able to see exactly where to make the required annotations using the inspection tool. After the development of this new approach to rejection sampling, the average trace length dropped to  $8.37$, with the longest registered trace being of length $42$. This made it possible to train the inference network and produce initial inference results for this particular physics problem.

\textbf{Initial physics results:} Using the setup described above, we were able to compute posterior distributions for $p_x, p_y, p_z, \texttt{channel}$ for various simulated observations with known ground truth. The discrete variable \texttt{channel} has a known prior distribution given by the branching ratio of the $\tau$ into $38$ possible decay channels~\cite{patrignani2016review}. While the \SHERPA{} code has been instrumented and inference is technically possible, at the time of this writing, we have not yet prepared an efficient similarity kernel used in ABC. The choice of this kernel is critical for balancing the efficiency of the posterior sampling and the quality of the approximate inference. With our initial similarity kernel we achieve accuracies  for the most frequent $\tau$ decay channels in the range of  $\approx$60-90\%. Therefore our further work aims to improve the similarity kernel used in ABC and improve the accuracy of the inference.



\textbf{Future work:} A common challenge for inference in particle physics is a large dynamic range of frequencies for various outcomes. For instance, some decay channels are much significantly more probable than the others, and the prior distribution for the momentum is often steeply falling. 
For this reason, the proposal parameters and the quality of the inference are better for observations with high likelihood.
 On the other hand, events with low likelihood (e.g., decay channels with small branching ratios) are less likely to be generated during training and the quality of the inference suffers. We are investigating techniques for automatically adjusting the measure during training to favor these unlikely outcomes, a feature that we name as ``prior inflation''.



The graph describing the probabilistic structure of the simulation that is generated by our inspection tool (Figure~\ref{fig:rejection_sampling}, a) constitutes an interesting candidate for learning a structured proxy (or surrogate) for the generative model itself. We are considering using such learned proxies for speeding up training and inference in models where the execution of the simulator code takes a significant amount of time and resources, such as the simulation of particle detectors in \GEANT.


In conclusion, this work provides an important first-step towards implementing a probabilistic programming approach in the physical sciences, which has considerable promise to leverage existing simulation software in order to provide tractable inference with a deeper level of interpretation than is possible with current analysis methods.

\section*{Acknowledgments}
Lezcano Casado is supported by an Oxford-James Martin Graduate Scholarship and a La Caixa Postgraduate Fellowship. Baydin and Wood are supported under DARPA PPAML through the U.S. AFRL under Cooperative Agreement FA8750-14-2-0006, Sub Award number 61160290-111668. Martínez Rubio is supported by Intel BDC / LBNL Physics Graduate Studentship. Cranmer, Louppe, and Heinrich are supported through NSF ACI-1450310, PHY-1505463, PHY- 1205376, and the Moore-Sloan Data Science Environment at NYU. This research used resources of the National Energy Research Scientific Computing Center (NERSC), a DOE Office of Science User Facility supported by the Office of Science of the U.S. Department of Energy under Contract No. DE-AC02-05CH11231.

\bibliography{refs}        
\bibliographystyle{plainnat}  
\end{document}